Aurélien Bénel, Joris Falip, Philippe Lacour

# “Quand Abel tue Caïn” : Ce qui échappe à la traduction automatique

*Although he emphasized that it is not yet possible "to insert a Russian book at one end and come out with an English book at the other," Doctor Dostert predicted that "five, perhaps three years hence, interlingual meaning conversion by electronic process in important functional areas of several languages may well be an accomplished fact.* (IBM, *communiqué de presse du 8 janvier 1954*[1]).

Des premiers objectifs du programme scientifique de l’Intelligence artificielle[2], aux récents succès de *DeepL* ou de *Google Translate*, la traduction a toujours occupé une place de choix dans le champ de l’Intelligence artificielle, peut-être parce qu’elle illustre à elle seule les deux significations du terme “intelligence” : celle de l’espion et celle de l’interprète ; intelligence de ce qui a une valeur stratégique ou économique chez l’étranger[3] et “lecture entre les lignes” des trésors d’une culture pour les transmettre à une autre[4].

Nous avons choisi d’aborder au travers d’une étude de cas la question de ce qui échappe à la traduction par une intelligence artificielle. L'intérêt est multiple. Il s’agit d’abord pour nous, de dépasser l’incommensurabilité de nos systèmes théoriques respectifs (épistémologie des SHS, ingénierie des connaissances, apprentissage automatique) en se donnant un objet partagé[5]. Il s’agit également d’identifier les situations où la traduction serait fautive. Un informaticien consciencieux, qu’il soit ingénieur ou chercheur, ne peut que prendre au sérieux des situations considérées comme des *bugs* reproductibles ou des réfutations du modèle entraîné.

Le cas que nous allons examiner est celui du récit de Caïn et Abel. Parmi les “grands textes de l’humanité”, ce récit, traduit dans toutes les langues et à toutes les époques (sans parler de ses nombreuses “traductions artistiques”), est intimement lié dans sa forme et ses lectures à une riche tradition interprétative, à une véritable herméneutique. Cette tradition critique, qui en un sens procède par “désacralisation” du texte, donne naissance aujourd’hui à une lecture très anthropologique[6] de ce récit de l’homme qui massacre son frère par jalousie, sujet d’une actualité sans cesse renouvelée… Par ailleurs, il n'aura échappé à personne que ce récit, par son ancienneté et son style, devrait *a priori* poser des difficultés aux traducteurs automatiques. Comme nous l'écrivions plus haut, personne ne sera surpris de la démarche qui consiste à discuter des cas qui semblent réfuter le modèle plutôt que de ceux qui semblent le corroborer. Qu'apprendrions-nous de l'étude des traductions censées être parfaitement automatisables comme celles des modes d'emploi[7] et de la documentation technique[8] ?

---

[1] Disponible sur : https://www.ibm.com/ibm/history/exhibits/701/701_translator.html

[2] John McDaniel, Translation by machine, *New Scientist*, vol. 19, issue 353, august 1963. p. 400–401.

[3] Michael D. Gordin, The Dostoevsky Machine in Georgetown: scientific translation in the Cold War, *Annals of Science, vol. 73*, issue 2, 2016. p.208–223.

[4] François Rastier, Communication ou transmission ?, *Césure 8*, 1995, p. 151–195.

[5] Aurélien Bénel. Quelle interdisciplinarité pour les « humanités numériques » ?. *Les Cahiers du numérique* 10 (4), Lavoisier, 2014, p.103-132. ⟨10.3166/lcn.10.4.103-132⟩.

[6] Delphine Horvilleur, *Comment les rabbins font les enfants : Sexe, transmission, identité dans le judaïsme*, première partie : “Fabrique du parent”, chapitre : “La faute à papa-maman”, Grasset & Fasquelle, 2015.

Nous commencerons notre étude par une observation des couples de langues proposés pour le cas qui nous occupe dans les interfaces des services les plus connus de traduction automatique. En faisant cela, nous en apprendrons un peu plus sur la stratégie des entreprises qui portent ces services, sur leur manière d'entraîner leurs modèles et sur les conséquences que cela peut avoir pour les usagers. Dans un second temps, nous lirons avec attention les résultats de ces services de traduction automatique et établirons une première caractérisation, voire une typologie, des erreurs rencontrées. Dans un troisième temps, nous comparerons entre elles des traductions contemporaines (notamment celle de Delphine Horvilleur, évoquée plus haut) et tâcherons d'en saisir les différences et l'apport singulier de chacune. Ainsi, nous pourrons tenter d'approcher le but poursuivi par les traducteurs de "textes culturels" (anciens, philosophiques ou littéraires) et ainsi de compléter ou de réviser la "théorie de la traduction" nécessaire à son instrumentation.

# Couples de langues : Premiers constats

Pour cette étude, nous avons comparé les textes produits par des outils de traduction automatique accessibles au grand public : d'un côté *Google Translate*, disponible dans 109 langues à l'heure actuelle, sûrement l'outil de traduction le plus utilisé au monde[9]. Lancé il y a maintenant plus de 16 ans, *Google Translate* a bénéficié des avancées de l'apprentissage automatique, notamment avec le passage à un moteur basé sur le *deep learning* en 2016, améliorant toujours ses traductions et augmentant le nombre de langues acceptées[10]. L'autre outil faisant l'objet de notre étude est *DeepL*, moteur de traduction européen et principal concurrent de *Google Translate*. *Deepl*, propriété de l'entreprise produisant la mémoire de traduction[11] *Linguee*, est notamment apprécié pour les technologies de traduction nouvelles qu'il a introduites dans le domaine de la traduction automatisée lors de son arrivée sur le marché il y a bientôt 5 ans.

---

[7] Notons toutefois que même les modes d'emploi et les étiquettes sur les produits, malgré leur simplicité, ne sont pas sans poser des difficultés aux traducteurs automatiques. Des groupes sur les réseaux sociaux (notamment "Traductions de m***" sur Facebook) en recensent d'ailleurs les meilleures perles.

[8] Annexe V du rapport de l'ALPAC, *Language and machines: Computers in translation and linguistics*, 1966

[9] *Google Translate* a été récemment installé sur plus d'un million de téléphones, auxquels s'ajoutent les utilisateurs du site web et de l'API. *https://blog.google/products/translate/one-billion-installs*

[10] Comme l'indique ce panorama des évolutions récentes de *Google Translate* https://ai.googleblog.com/2020/06/recent-advances-in-google-translate.html

[11] Mémoire de traduction : Base de données dans laquelle les traducteurs stockent des "segments" traduits de manière à être réutilisés plus tard.
Aurélien Bénel, Philippe Lacour. Towards a Participative Platform for Cultural Texts Translators. El Morr, Christo; Maret, Pierre (Eds.), *Virtual community building and the information society : Current and future directions*. IGI Global, 2012. pp.153-162.

A ces deux outils, nous avons soumis comme cas d'étude deux versions du récit de Caïn et Abel : l'une en hébreu (version dite “massorétique”[12]) et la seconde en grec ancien (version dite “de la Septante”[13]). Les résultats sont variés et dépendent aussi bien des langues que de l'outil utilisé. Avant d'examiner les traductions en français, nous pouvons dresser quelques constats préliminaires, à commencer par une différence évidente : le choix des langues prises en charge par chaque service. Si *DeepL* accepte bien le grec ancien, il s'en accommode en considérant étonnamment le texte source comme étant analogue à du grec moderne. Par contre *DeepL* ne parvient pas à identifier la langue du texte hébreu qu'il détecte comme de l'anglais, il propose donc comme traduction française une recopie du texte source qu'il considère comme un verbatim intraduisible. Nous avons une situation opposée pour *Google Translate*, qui s'accommode correctement de l'hébreu pour lequel il propose une traduction française. L'outil de *Google* reconnaît bien le texte en grec ancien, mais ne propose toutefois en guise de traduction française qu'un texte parcellaire en grec moderne. Nous pouvons constater immédiatement l'impact des choix opérés par les deux entreprises vis-à-vis des langues qu'elles proposent au sein de leur outil respectif, chacune décidant des langues justifiant l'investissement financier nécessaire à leur intégration à l'outil de traduction. Dans le cas de *DeepL* nous pouvons par exemple supposer que la prise en charge du grec mais pas de l'hébreu résulte d'une décision de réutiliser les données de *Linguee* qui, parce que construites notamment en s'appuyant sur les textes du parlement européen, possèdent de très vastes corpus en grec mais pas un seul texte en hébreu. Notons aussi que les moteurs de traduction semblent, tout comme la nature “aristotélicienne”, avoir horreur du vide. Les deux outils refusent de rendre copie blanche même pour les langues qui ne sont pas traitées, quitte à n'effectuer alors pour seule proposition qu'une simple recopie de l'entrée qui leur a été fournie.

Nous pourrions opposer à ces échecs qu'ils ne sont pas tant des erreurs que des étapes de l'apprentissage automatique et n'invalident pas l'outil, qu'il suffirait d'entraîner sur de nouveaux jeux de données (en hébreu pour *DeepL* et en grec pour *Google Translate*) pour combler ces manques.

Au-delà de ce point, nous touchons aussi ici aux limites de méthodes de traduction qui reposent sur l'usage massif de données, où le besoin exponentiel de données d'entraînement est confronté à la réalité financière et aux limites matérielles régissant l'agrégation de ces données. Bien que le processus d'apprentissage automatique puisse être poursuivi pour en améliorer la complétude, des entreprises de l'ampleur de *DeepL* ou *Google* ne trouvent pas les ressources (ou plutôt le retour sur investissement) pour traduire des langues plus rares. Ce coût du travail humain, nécessaire à l'élaboration des jeux de données, est aussi ce qui réserve le domaine de la traduction automatique aux plus grands acteurs du numérique et écarte une grande partie du paysage académique.

Dans les cas où l'amélioration du modèle n'est pas empêchée par les critères évoqués précédemment, compléter le jeu de données pour améliorer les performances de l'algorithme d'apprentissage pose une autre difficulté car cela résulte en un nouveau modèle. Plus le modèle entraîné est complexe et plus il devient délicat de s'assurer que ce nouveau modèle, normalement “renforcé” et amélioré par l'extension du jeu de données, ne cache pas une perte de performance sur les données précédentes ou un “surapprentissage” (*overfitting*) qui se ferait au détriment de l'adoption de cette nouvelle version de l‘outil.

[12] Codex de Leningrad, 1008.

[13] Codex Vaticanus et Codex Sinaiticus, IVe s.

En admettant que ces écueils soient évités, le mode d’apprentissage même de ces réseaux de neurones s’oppose au geste du traducteur car l’outil repose sur l’induction, qui tente de dériver par abstraction une vérité généralisée à partir d’une collection de cas particuliers, dans notre cas une règle de traduction. On s’éloigne donc de l’approche du traducteur pour qui chaque texte est traité dans toute sa singularité[14] et sa complexité.
En utilisant comme base d’apprentissage un ensemble de cas particuliers, l’intelligence artificielle s’éloigne déjà du singulier : le particulier en étant une simplification qui postule que chaque texte est une illustration particulière, une incarnation ponctuelle d’une règle universelle.

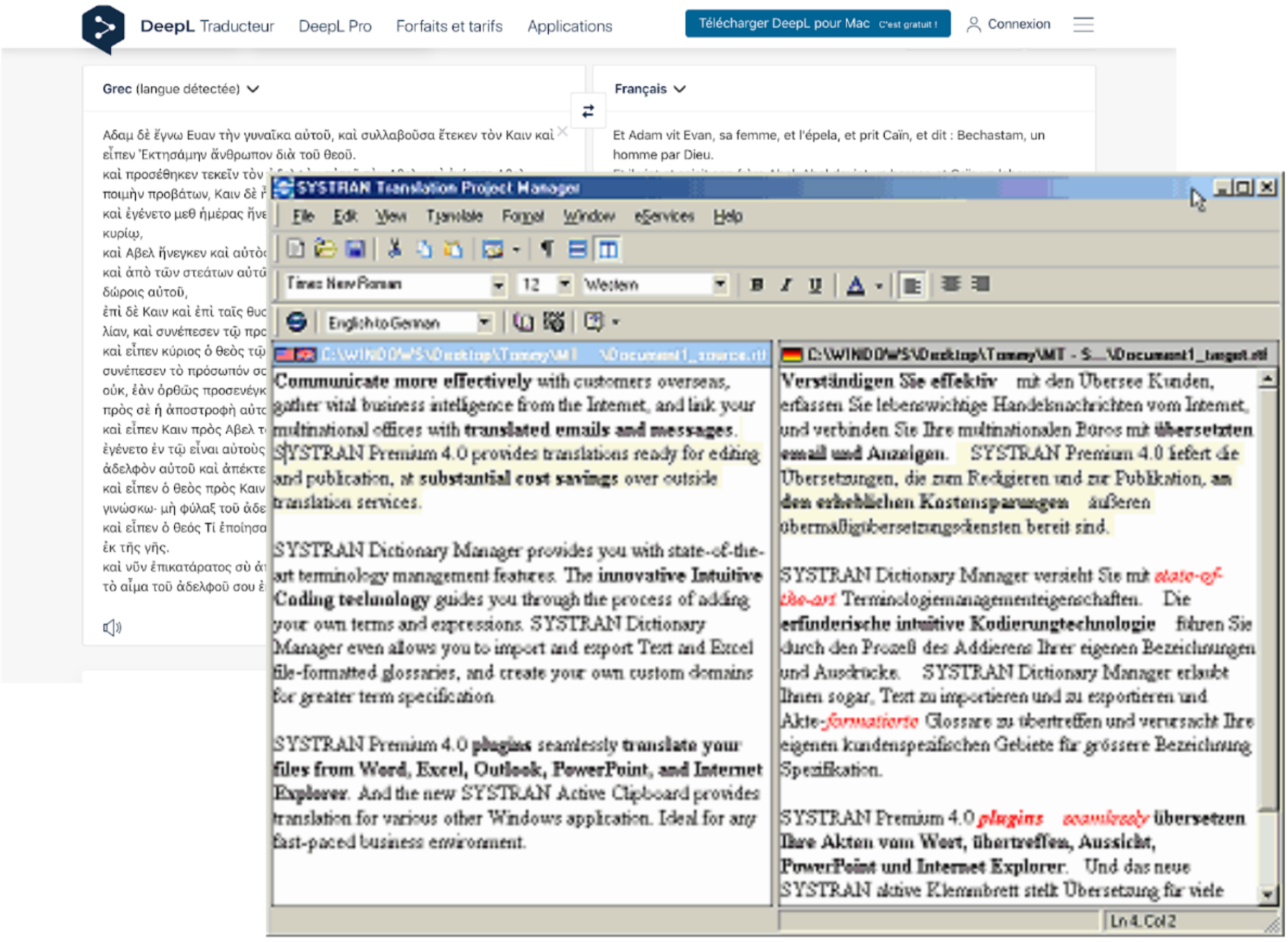

**Figure 1.** Copies d’écran des interfaces utilisateur à une vingtaine d’années d’intervalle des systèmes de traduction automatisée DeepL et Systran.

Si nous prenons du recul pour analyser non plus les données et inférences qui en sont faites mais le modèle d’interaction avec ces outils, nous remarquons qu’il est inchangé depuis au moins vingt ans (cf. Figure 1). Cette manière de présenter le processus de traduction (avec une entrée dans une langue, une sortie dans une autre et un traitement automatique permettant de passer de l’un à l’autre) s’explique par les présupposés à l’origine de la traduction automatique. En effet, au début de la guerre froide, forte des succès du déchiffrement automatisé des messages ennemis lors de la seconde guerre mondiale, le mathématicien Warren Weaver[15] de même que la R&D d’IBM imaginent compléter le décryptage des messages russes par une traduction vers l’anglais. La traduction est donc vue comme une simple extension de la cryptanalyse et donc comme un traitement de même nature.

Fruits d’un véritable impensé, les systèmes s’appliquent à traduire par une bijection selon laquelle tout texte dans une langue source (à gauche dans les outils de traduction) possèderait un unique texte équivalent dans la langue cible (colonne de droite de chacun de ces outils), il suffirait donc d'entraîner l'algorithme à reconstituer au mieux ces couples de textes supposés identiques, quitte à passer systématiquement par une langue pivot dans l'idée que tout texte possèderait un homologue parfait dans toutes les langues.

---

[14] Le singulier désigne ce qui est unique tandis que le particulier vise ce qui vaut pour un certain nombre de cas (au prix d'une certaine mise en commun). La singularité à laquelle nous nous référons est celle du texte, or l'universel n'admet pas d'exception tandis que le général si.
Philippe Lacour, *La raison au singulier : Réflexions sur l’épistémologie de Jean-Claude Passeron*, Presses Universitaires de Nanterre, 2020.
[15] Warren Weaver. *Translation Memorandum*. 1949.

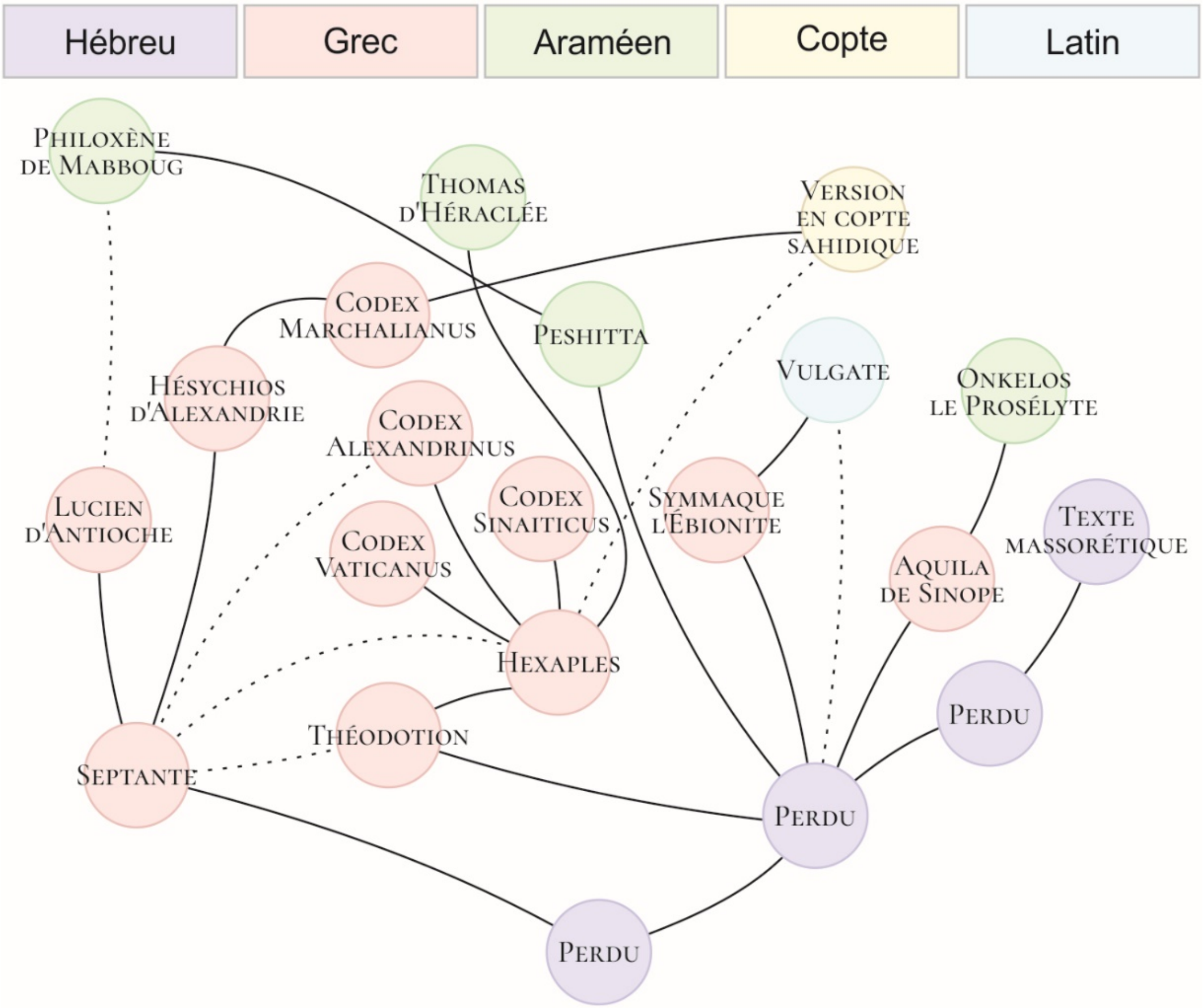

**Figure 2.** Arbre des traductions de la bible hébraïque (dont le récit de Caïn et Abel est tiré).

À l'inverse de cette simplification d'ingénieur, l'acte de traduction emporte avec lui tout un contexte du texte singulier dont le traducteur transforme la signification en une signification analogue dans plusieurs autres langues tout en l'ancrant dans un genre mais aussi une époque, les traductions évoluant avec le temps et la société. Loin de la bijection, le traducteur dans son travail s'appuie en effet souvent sur la connaissance de plusieurs langues et sur l'utilisation de plusieurs traductions concurrentes pour produire le texte souhaité[16]. Le cas de la bible hébraïque, dont notre récit est tiré, est d'ailleurs emblématique de cette complexité. L'arbre des traductions anciennes de la bible (cf. Figure 2) fait figurer les versions attestées par des manuscrits, d'autres de manière plus indirecte par des références dans d'autres textes anciens, d'autres enfin ne sont que des hypothèses de travail des philologues pour expliquer les ressemblances entre manuscrits. On y découvre en particulier :

- une traduction (comme celle de Lucien ou d'Hésychius) qui en améliore une plus ancienne (la Septante) dans la même langue (ici le grec),
- une autre (celle de Philoxène) qui pour en améliorer une précédente (la Peshitta) dans une langue (ici le syriaque) a recours à une version plus ancienne dans une autre langue (celle de Lucien en grec),
- un travail érudit sur le sens (ici la vocalisation d'un texte transmis sous forme de consonnes pour produire le "texte massorétique"), analogue à celui de la traduction même si le texte produit est dans la même langue que le texte original,
- une traduction radicalement concurrente à une autre (celle de Theodotion par rapport à la Septante pour le grec),

[16] Aurélien Bénel, Philippe Lacour, *op. cit.*

- une compilation des différentes traductions (grecques) dans un même ouvrage (hexaples),
- une traduction dans une nouvelle langue (le sahidique) à partir de deux versions concurrentes dans une langue (ici le grec),
- une traduction dans une nouvelle langue (la vulgate, en latin) à partir de deux versions dans des langues différentes (ici le grec et l'hébreu).

Au-delà d'une meilleure perception du geste du traducteur, beaucoup moins spontané, plus érudit et ancré dans la comparaison des différences entre singularités, cette généalogie nous renseigne aussi sur les attentes des lecteurs : le moins que l'on puisse dire, c'est que, pour le texte qui nous occupe, ils ne se sont pas contentés d'une traduction approximative dans leur langue…

# De part et d'autre de la boîte noire

Des travaux comme les nôtres se heurtent à une difficulté spécifique au champ : les outils de traduction automatique sont de parfaites boîtes noires. Cette opacité est double. Elle concerne d'abord le processus de constitution des données d'entraînement (repérage des sources de données, négociation des droits, numérisation éventuelle, prétraitement manuel ou automatique). En effet, comme tout processus d'apprentissage automatique, la constitution du jeu de données destiné à entraîner le modèle est une étape cruciale. Or, on comprend sans peine que les acteurs, dans un but de compétitivité, ne souhaitent pas partager ce qui s'apparente à un secret industriel. Mais cette opacité est également inhérente aux méthodes d'apprentissage utilisées : basées sur les réseaux de neurones, elles sont incompatibles avec tout effort de rétro-ingénierie visant à déterminer comment une traduction a été produite. Il nous sera donc impossible (tout au moins de manière directe) d'expliquer précisément comment la constitution des données et la structure du modèle entraîné impactent la qualité de la traduction. Par contre, ces services étant mis gratuitement à la disposition de tous, ils permettent des tests dits "en boîte noire" : nous pouvons, en choisissant les entrées (*input*), étudier les sorties (*output*).

Comme nous l'avons vu précédemment, pour le récit qui nous occupe, seules deux traductions automatiques seront utilisables : celle à partir du texte grec pour *DeepL* (cf. Figure 3) et celle à partir du texte hébreu pour *Google Translate* (cf. Figure 4).

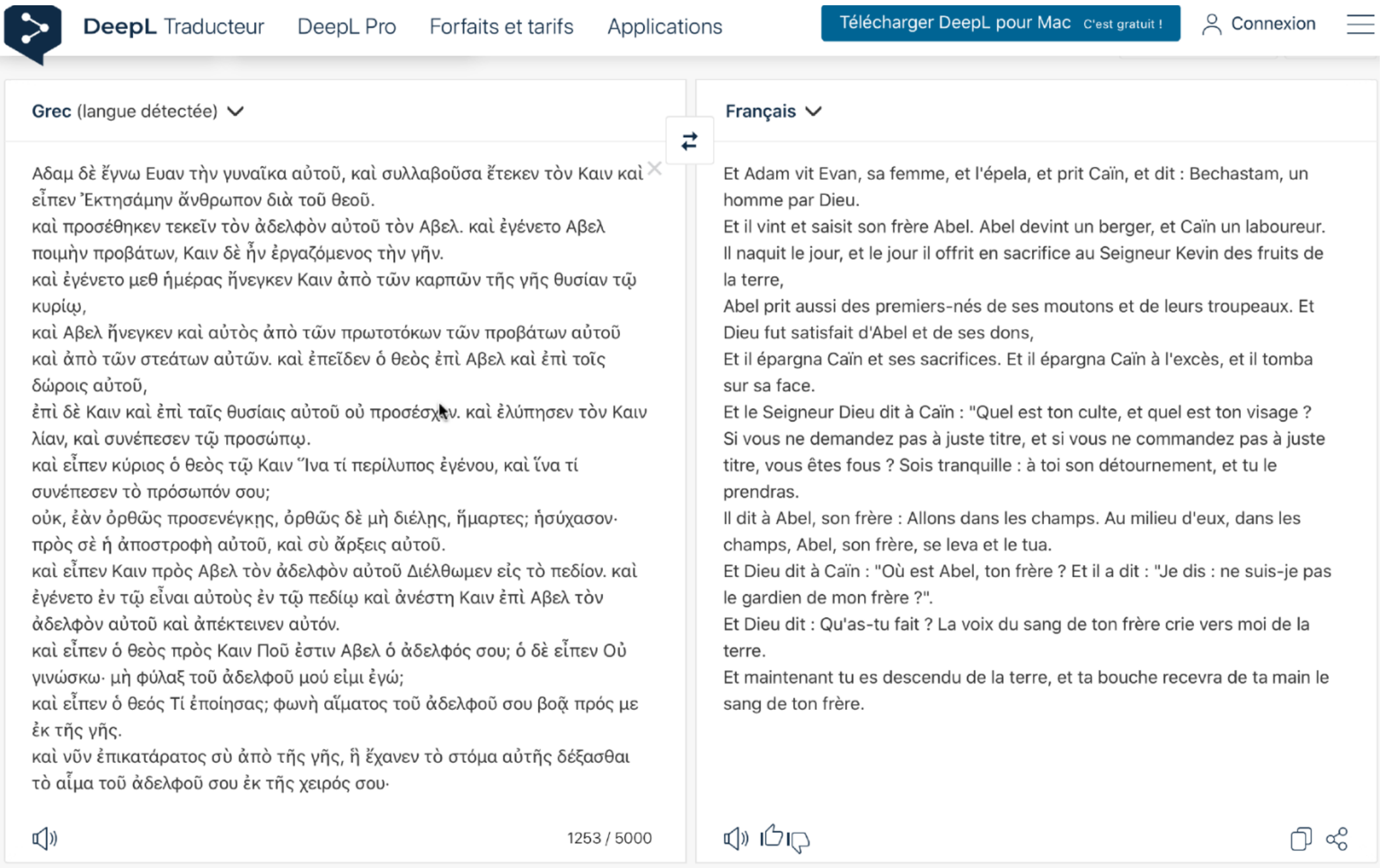

**Figure 3.** Traduction automatique en langue française du récit dans sa version dite "de la Septante" (Copie d'écran de *DeepL*).

Étudions la traduction bloc après bloc (verset après verset).

(1) "Et Adam vit Evan"... Et dès les premiers mots, *DeepL* bute sur l'incommensurable variabilité des formes dans une langue où même les noms propres se déclinent. "Il l'épela". Adam "épèle"-t-il Ève ? Ou ne serait-ce pas plutôt Ève qui "épèle" Caïn ? La résolution anaphorique des pronoms – problème classique et redoutable en traitement automatique des langues - est ici d'autant plus difficile qu'en grec les pronoms sont souvent implicites. Le choix du terme "épeler", bien qu'un peu surréaliste, n'est pas absurde. Il est bien question d'une action qui se fait petit bout par petit bout, mais qui serait plutôt de l'ordre de la gestation. Le terme "Bechastam", lui, est tout à fait mystérieux. Comme il ne s'agit ni d'un terme français ni de la translittération du grec, on peut faire l'hypothèse qu'un des couples de textes utilisés lors de l'apprentissage contenait une coquille (peut-être issue d'une autre langue). Hélas le modèle étant une boîte noire, cette bizarrerie restera inexplicable.

(2) "Il vint et saisit son frère". Cette image surprenante d'un nouveau-né empoignant son jumeau, même si elle existe dans le même livre une vingtaine de chapitres plus loin, ne doit sa présence ici qu'à cause d'une erreur de résolution de l'anaphore pronominale, et une maladresse dans le choix de traduction du verbe ("produire", "donner la vie").

(3) La formule "Le Seigneur Kevin", digne des Monty Python, est intéressante autant par son étrangeté que par le fait qu'elle ait été corrigée depuis. Il s'agit maintenant de "l'Éternel". Or ce dernier terme est la substitution du nom imprononçable (YHWH) dans les traductions réalisées à partir de l'hébreu. Dans le grec, au contraire, on lui a substitué "Kyrios" ("Le Seigneur", littéralement "le monsieur"). Autrement dit, cette "correction" de la traduction s'est faite en ajoutant aux données d'apprentissage des traductions françaises de la bible, y compris des traductions à partir de l'hébreu en faisant *comme si* c'étaient des traductions à partir du grec. Générer artificiellement des données à partir des données existantes n'est pas rare en apprentissage automatique (où les données coûtent si cher). Cependant cette pratique se heurte ici à un point fondamental : en faisant ainsi, les employés de *DeepL* ont fait comme s'il existait une classe correspondant à l'œuvre dans toutes ses traductions. La sémiotique (ou la sémantique interprétative) nous apprend au contraire que le sens n'est pas dans l'appartenance à une classe, mais dans la différence entre deux singularités[17] [18]. Ou pour le dire plus simplement, le geste de traduction consiste en une opération entre textes concrets, et non en une extraction d'un exemplaire à partir d'un sac abstrait qui engloberait toutes les traductions en en faisant une abstraction.

(4-6) Les erreurs des trois versets suivants sont analogues à celles déjà vues (anaphores pronominales, mauvais choix parmi les différentes significations des termes).

(7) "Si vous ne demandez pas à juste titre, et si vous ne commandez pas à juste titre, vous êtes fous ? Sois tranquille : à toi son détournement et tu le prendras". La traduction est pour le moins confuse… Ceci est d'autant plus regrettable que sa position dans le récit et son style, effectivement un peu différent, semble en faire une sorte de parole de sagesse, de dicton qui pourrait éclairer le sens du texte tout entier. Inversement, c'est aussi le texte qui permettrait d'éclairer ce passage, visiblement assez obscur. Nous touchons ici un autre point fondamental appelé "cercle herméneutique"[19] : quand il s'agit d'interpréter un texte, la compréhension globale d'un texte influe sur la compréhension locale autant que la compréhension locale influe sur la compréhension globale. Ce point essentiel de la théorie de l'interprétation entre en contradiction frontale avec l'hypothèse, difficilement évitable en traitement automatique des langues, selon laquelle le sens d'un texte pourrait être obtenu par composition de ses parties.

---

[17] François Rastier, Sens et signification, In : *Protée, printemps 1998*. p.7-18.
[18] Philippe Lacour. L'oubli de la sémantique dans le programme cognitiviste : réflexions sur l'œuvre de François Rastier. *Texto ! décembre 2004*
[19] Denis Thouard, Servanne Jollivet, "Cercle herméneutique", in *L'interprétation. Un dictionnaire philosophique*, Paris, Vrin, p. 75-80, 2015.

(8) “Abel, son frère, se leva et le tua”. L’erreur de résolution de l’anaphore pronominale ne pouvait être plus grave pour la compréhension du sens du texte : la victime devient assassin et l’assassin victime !

(9) “Ne suis-je pas le gardien de mon frère ?”. On s’attendrait au contraire à ce que Caïn nie ou interroge le fait d’être “le gardien de son frère”.

Les deux dernières erreurs pourraient sans doute être qualifiées de “contre-sens”. Or la pertinence de ce concept habituel de la théorie de la traduction est aujourd’hui contestée. Des erreurs aussi graves, en effet, ne se produiraient jamais dans les traductions professionnelles (humaines)[20].

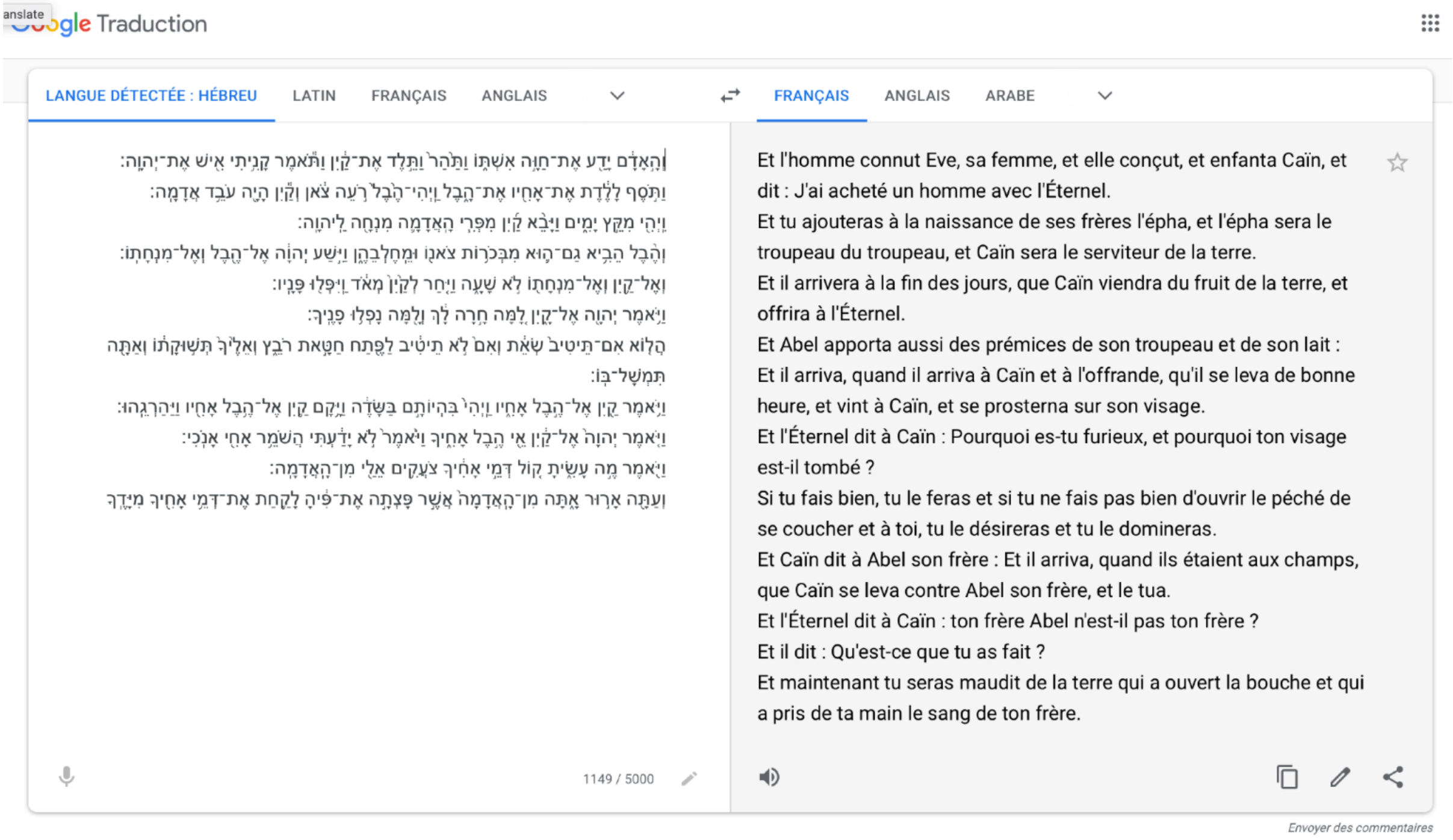

**Figure 4.** Traduction automatique en langue française du récit dans sa version dite “Massorétique” (Copie d’écran de *Google Translate*).

Concernant la seconde traduction automatique, celle de *Google Translate* à partir de l’hébreu, la densité d’erreurs est plus faible. Attardons-nous sur les versets dont la traduction pose cependant problème.

(1) On pourrait être étonné du choix du verbe dans “J’ai acheté un homme”. Mais s’il y a une erreur, il s’agit d’une erreur de nuance, car, comme on le verra plus tard, le terme original porte effectivement une connotation de cet ordre. Le degré est cependant trop fort par rapport au contexte.

(3) “[Caïn] viendra du fruit de la terre” est syntaxiquement correct, presque poétique, mais n’a malheureusement que peu de rapport avec le texte original. Le fait que les traductions *ressemblent* de plus en plus à des textes corrects est paradoxalement un vrai problème. En effet, comme nous l’avons vu précédemment, par conception, ces outils donnent toujours une réponse même dans des situations où il serait plus ajusté de donner une réponse vide ou un message d’erreur. Le passage d’une IA symbolique à une IA statistique a formidablement amélioré la ressemblance des phrases produites artificiellement avec des phrases naturelles. La contrepartie est que la détection par l’utilisateur des passages erronés (par exemple pour les corriger à la main) n’a jamais demandé autant d’effort et de temps.

(4) Le problème est similaire quand il est question des “prémices de son troupeau et de son lait”. Toute bucolique et rassurante que soit l’image, le choix du mot “lait” s’écarte de celui de l’ensemble des traducteurs qui y voient de la “graisse”. On peut penser que ce choix s’explique par une plus grande fréquence dans les corpus de la collocation de “troupeau” et de “lait” que celle de “troupeau” et de “graisse”. Là encore, il est possible que le seul fait que l’approche soit statistique vienne favoriser ce qui est plus “lisse”, plus attendu.

[20] Dussart, André. Faux sens, contresens, non-sens… un faux débat ?, *Meta 50(1)*, mars 2005, p. 107–119. <10.7202/010661ar>

(5) L’élément déclencheur du drame, n’en déplaise à *Google Translate* et aux Proustiens, n’est pas,que ‘[l’Éternel] se leva de bonne heure”, mais qu’il se tourna vers l’offrande d’Abel et non celle de Caïn.

(7) Dans le fameux “passage obscur” il est maintenant question “d’ouvrir le péché de se coucher” (!), avec par ailleurs les fonctions grammaticales des différentes parties qui semblent toutes fausses.

(9) “Ton frère Abel n’est-il pas ton frère ?”. À nouveau, le problème est plus grave quand la traduction est syntaxiquement correcte et même qu’elle a un sens (y compris en contexte), mais qu’elle n’a quasiment rien à voir avec ce que dit le texte.

Pour finir, on peut noter que certaines phrases de ces traductions sont exemptes d'erreurs. D’aucuns pourraient espérer que dans un futur plus ou moins proche, cela devienne le cas du texte entier… Mais dans ce cas précis qu’auraient réellement gagné ses lecteurs ? De nombreuses traductions existent déjà de ce texte, et certaines d’une grande finesse…

Il est temps maintenant de voir ce qui distingue une traduction “humaine” d’une traduction automatique. Au lieu de chercher à instancier des tendances, à ranger l’inconnu dans les “boîtes” de ce qui est connu, la traduction humaine – quand elle est de qualité – cherche ce qui est singulier, ce qui est différent : d’une langue à l’autre, d’un texte à un autre, d’un passage à l’autre. Elle met l’accent sur la singularité du sens, l’originalité du contexte, l’unicité de la signification visée, et sur l’opération d’interprétation qui vise à inventer un *équivalent sans identité*[21].

# D’une traduction à l’autre : différences et singularité

Troisième étape de notre parcours et nouvel artefact à étudier : quatre traductions de notre récit, produites par des contemporains, en langue française, en autant de colonnes, alignées verset par verset (cf. Figure 5).

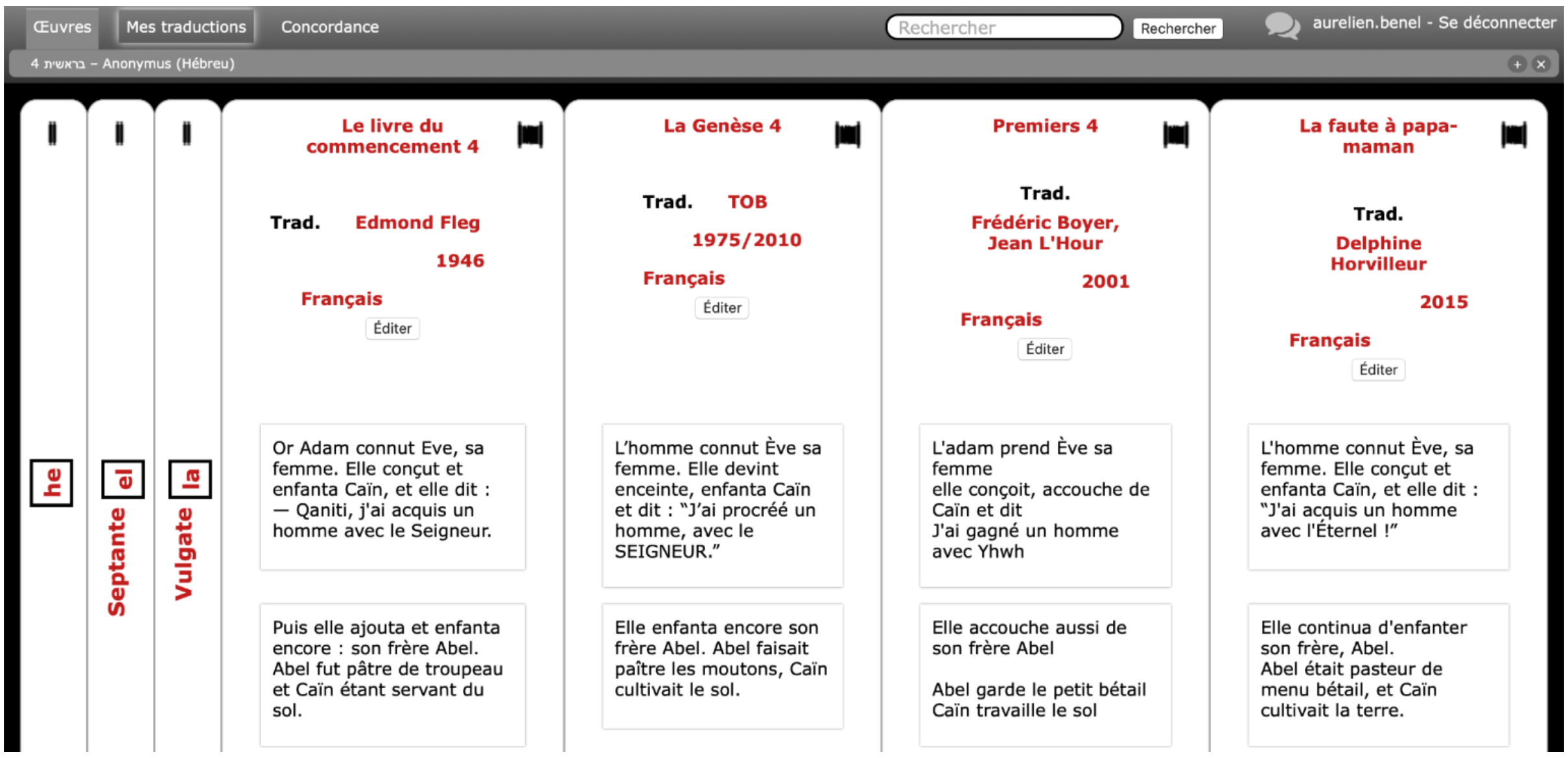

[21] Paul Ricoeur, *Sur la traduction*, Paris Bayard, 2004.

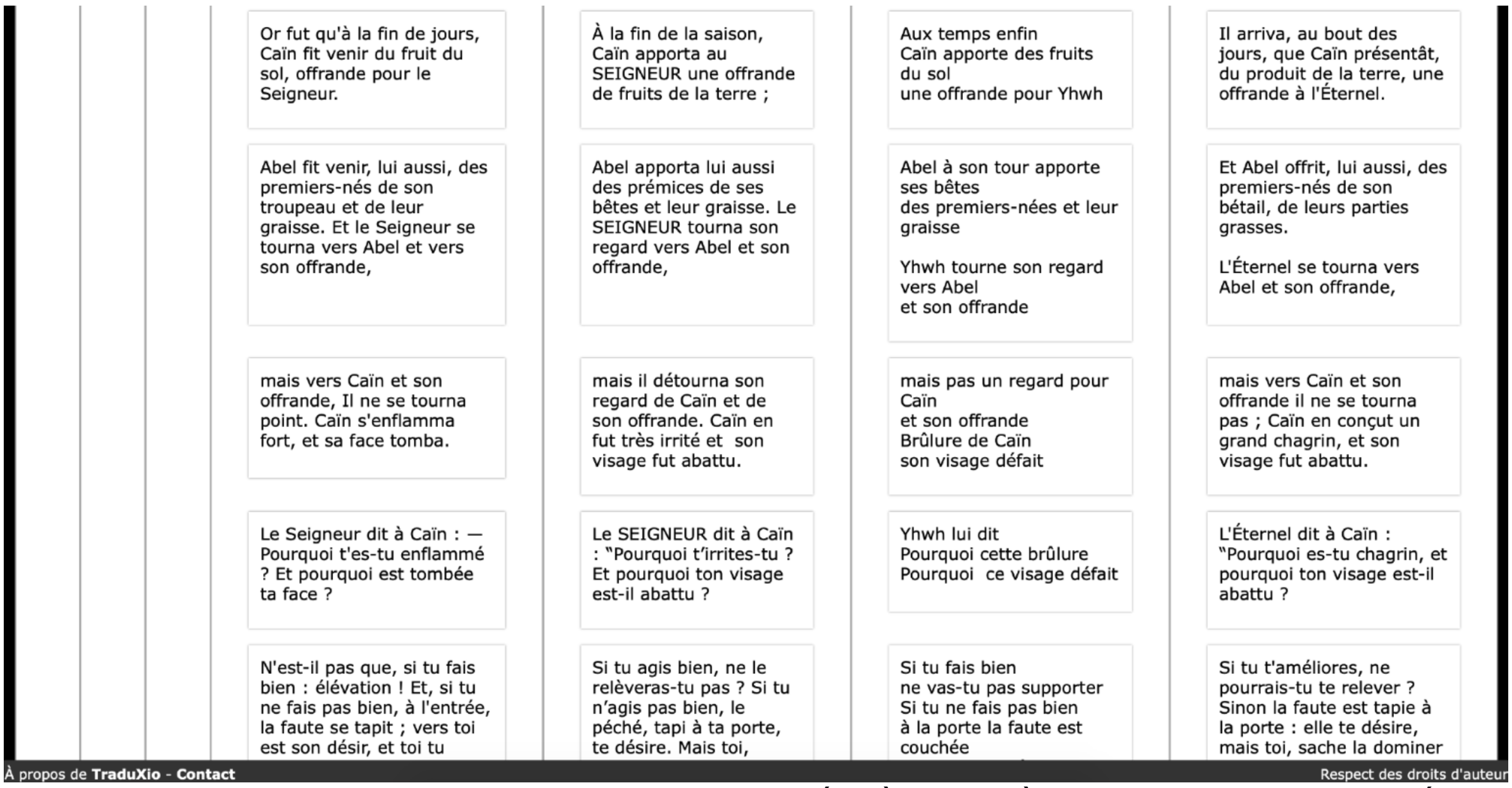

**Figure 5** : Traductions françaises "concurrentes" du récit à la manière des *Hexaples* (copie d'écran de *TraduXio*).

Le fait que ces textes soient lus dans un logiciel (*TraduXio*[22]) est ici un prétexte : prétexte à faire remarquer que, contrairement à un certain mythe de l'informatique, celle qui a changé profondément notre vie quotidienne et notre travail est plus liée à un changement de support[23] qu'à une réelle automatisation ; prétexte également à souligner que, si l'informatique est affaire de support et de forme visuelle, il peut être pertinent de nourrir la conception des logiciels de la riche tradition des usagers auxquels ils sont destinés (ici, la présentation en colonnes veut rappeler celle des *Hexaples* mise au point à Alexandrie au III^e^ s. comme support au débat entre juifs et chrétiens autour des différentes traductions en grec de la Bible hébraïque[24]).

La simple existence de ces quatre traductions, dans la même langue, en moins de quatre-vingts ans, nous pose la question de ce qui pousse à traduire et à traduire encore ces textes. Si chaque génération souhaite avoir sa traduction, c'est peut-être parce qu'il s'agit de *compréhension*. Ce "prendre avec" suppose deux langues, deux époques, deux cultures. Or les unes et les autres évoluent : celles des traducteurs, bien-sûr (la langue mais surtout les questions qui traversent la société ne sont plus les mêmes aujourd'hui qu'après-guerre), mais celles des auteurs également dans la perception que l'on en a (grâce notamment aux avancées des travaux historico-critiques). Dès lors, nous l'avons compris, il serait illusoire de penser qu'un texte est traduit une fois pour toutes.

Ce travail de compréhension que requiert la traduction nous fournit une autre raison de traduire toujours et encore. Dans le cas de Delphine Horvilleur, son but n'est pas de faire une traduction de toute la Genèse encore moins de toute la Torah, mais juste de quelques passages qui lui semblent éclairants à propos d'une vision juive du sexe, de la transmission et de l'identité[25]. Pour le linguiste François Rastier, d'ailleurs, les meilleurs commentateurs d'un texte sont ceux qui l'ont traduit[26]. Essayons précisément de voir ce que la traductrice apporte à la compréhension du texte au-delà d'une traduction qui serait simplement "correcte".

---

[22] Philippe Lacour, Aurélien Bénel. TraduXio Project: Latest Upgrades and Feedback. *Journal of Data Mining and Digital Humanities*, Numéro spécial sur les humanités numériques, Episciences.org, 2021, ⟨10.46298/jdmdh.6733⟩, ⟨hal-02920044v3⟩.

[23] Des philosophes de l'informatique proposent de considérer l'intelligence artificielle elle-même (au moins dans sa tradition symbolique) comme une nouvelle étape dans l'évolution des supports, retracée par Jack Goody, démultipliant encore les dimensions de lecture (Bruno Bachimont, L'intelligence artificielle comme écriture dynamique : de la raison graphique à la raison computationnelle. In: Petitot, J, Fabbri, P. (Eds.) *Au nom du sens: Autour de l'œuvre d'Umberto Eco*, Grasset, 2000, pp. 290–319).

[24] Aurélien Bénel. *op. cit.*

[25] Delphine Horvilleur, *op. cit.*

(1) Les choix de traduction commencent dès le premier mot : comme dans la plupart des traductions contemporaines, la traductrice voit dans le mot "*adam*" non pas un nom propre, mais un nom commun ("le glaiseux"), qu'elle choisit de traduire par "l'homme" afin de faire ressortir le caractère anthropologique du récit. Comme d'autres également, elle choisit le verbe "acquis" pour expliciter l'étrangeté du lien entre Ève et le nouveau-né, ce terme un peu gênant qui devient son prénom : "Caïn". La traductrice redevient essayiste en notant dans son commentaire que tout se passe comme si Caïn appartenait à sa mère. Le père, lui, est doublement absent : absent du texte après sa courte participation et substitué par le transcendant dans la déclaration d'Ève à propos de l'"acquisition".

(2) "[Ève] continua d'enfanter"... Ce "frère", note Delphine Horvilleur, est le "petit supplément" de la "grandiose" première naissance, cet aîné présenté par sa mère à la fois comme sa possession et comme un demi-dieu. Pour ajouter à son caractère transparent, le cadet est nommé Abel, ce qui veut dire "buée"...

(3-6) Puis, quand vient "le bout des jours", Abel-la-buée imite Caïn-le-demi-dieu-de-sa-maman en faisant une offrande au transcendant qui incarne une certaine figure paternelle. Et là, sans aucune explication, ni raison apparente, la figure paternelle préfère le petit dernier. La traductrice choisit l'expression un peu décalée de "concevoir un grand chagrin"... Peut-être pour montrer le caractère enfantin et ordinaire de cette jalousie de l'homme pour son frère ? Peut-être aussi parce que cette expression lui permet dans le verset suivant d'utiliser la forme "être chagrin" : ce n'est plus juste une émotion, c'est tout son être qui en est transformé.

(7) Le passage obscur s'éclaire : contraste et choix possible entre le relèvement auquel l'homme est invité, après ce terrible abattement, et ce qui se "tapit" (à la fois, "bas", comme les "bas instincts", et menaçant, dangereux). La commentatrice qui s'est confrontée à la traduction note une intraduisible incohérence dans les temps des verbes : "Quand tu t'amélioreras (futur), tu t'es relevé (passé)". Elle en retient pour son commentaire qu'il s'agirait d'une sorte de définition biblique de la résilience (du rebond) : "s'élever au futur d'un passé qui abat".

On pourrait continuer la lecture de Delphine Horvilleur, mais le but ici était d'illustrer à la fois ce qui échappe à la traduction automatique (au-delà de ce qui empêche la traduction d'être "correcte") et les raisons pour lesquelles on traduit un texte, encore et toujours, à chaque génération, tout simplement parce que traduire est peut-être la meilleure manière de lire un texte en profondeur.

# Conclusion

Que peut-on conclure de notre itinéraire à travers les traductions automatiques et humaines du récit de Caïn et Abel ? Pour ce qui est de la traduction automatique elle-même, le premier constat est une amélioration indéniable, ces dernières années, de la qualité de la syntaxe et de la cohérence perçue des phrases générées par les outils de traduction automatique. Pour autant, et c'est là notre deuxième constat, les faux-sens, voire les contre-sens (comme celui que nous citons dans le titre) restent très nombreux. L'argument d'une amélioration prochaine voire imminente de ces points est peu convaincant puisque, pour la plupart, ces problèmes (notamment la résolution anaphorique des pronoms) sont déjà au centre de l'attention du champ du traitement automatique des langues depuis plusieurs décennies. Ces deux constats entremêlés font de certains passages traduits des éléments *ressemblant* tellement à un texte en français, qu'il était difficile pour nous d'admettre que ces passages n'ont que très peu à voir avec le sens du texte original.

---

[26] François Rastier, La traduction: Interprétation et genèse du sens, Lederer, M. (Ed.), *Le Sens en traduction*, Cahiers Champollion, Minard, 2006.

Du côté de l’usage, même conscients de ces graves erreurs, certains défendront la possibilité offerte par ces outils de se faire une idée du *sujet* d’un texte, ce qui finalement nous ramène à la visée originelle de la traduction automatique en pleine guerre froide, visant à décrypter les communications. D’autres, revendiqueront l'intérêt de la traduction automatique pour produire une version “martyr” en vue d’une traduction manuelle. Cependant, en raison de l’absence d’indices sur la qualité des différents passages traduits de manière automatique[27] et même d’une perception faussée de la qualité, il n’est pas évident que la correction de cette version intermédiaire soit moins coûteuse et produise de meilleurs résultats que la traduction manuelle directe.

Notre troisième constat concerne les conséquences sur les services de traduction automatique des choix effectués par les entreprises qui les fournissent : choix des langues gérées, des corpus d’apprentissage, de ce qui peut être considéré comme une traduction (menant par exemple, comme nous l’avons vu, à ce que deux traductions de versions différentes de la même œuvre soient considérées comme des traductions l’une de l’autre). Ces choix, probablement dictés par le marketing, laissent en dehors de leur périmètre des pans entiers de la connaissance, de la culture et des pratiques humaines. Par ailleurs, le coût très élevé des moyens informatiques et du travail humain nécessaires à la mise en place d’un tel service empêche tout acteur public d’offrir des services concurrents pour répondre à ces besoins considérés comme “de niche”.

D’un point de vue plus épistémologique, on pourrait nous objecter qu’il est de toute manière trop tard pour résister à un mouvement de “mathématisation du réel” ou, plus spécifiquement, d’”apprivoisement du probable” qui sont à l’œuvre respectivement depuis le XVIIe et le XVIIIe siècles[28]. Notre souci n’est pas tant de s’opposer à toute modélisation que de veiller à ne pas perdre l’essentiel au cours de ce processus. Comme nous l’avons vu, l’analogie au décryptage d’un message codé, mobilisée dans ces outils, est extrêmement naïve en comparaison du raffinement des pratiques attestées et des théories de la traduction, notamment vis-à-vis de la prise en compte du singulier dans la tradition herméneutique.

Dès lors nous aimerions ouvrir deux perspectives. La première est peut-être déjà à l’œuvre dans cet ouvrage collectif. Nous pourrions la formuler ainsi : il est temps de prendre au sérieux l’*Intelligence artificielle*, non comme un ensemble de techniques, mais comme un programme scientifique, celui de la conférence de Dartmouth en 1956 qui imaginait une communauté interdisciplinaire qui se consacrerait à l’étude de l’intelligence humaine pour en développer une meilleure compréhension[29]. Conséquence de ce retour aux sources de l’intelligence artificielle, la seconde perspective serait d’accorder une considération au moins aussi grande à l’outillage multiséculaire du travail intellectuel qu'à l’apprentissage automatique. Par exemple, en transposant les hexaples et les concordances dans un environnement participatif destiné aux traducteurs, un logiciel comme *TraduXio*[30] préfigure selon nous ce que ce genre d’approches apporte non seulement à l’assistance au travail intellectuel mais également à sa théorisation.

---

[27] À la différence d’autres traitements statistiques comme, par exemple, la reconnaissance optique de caractères.

[28] Giorgio Israël, *La mathématisation du réel*, Paris, Seuil, 1996.
Ian Hacking, *L’Émergence de la probabilité*, Paris, Seuil, 2002 ; *The Taming of Chance*, Cambridge, Cambridge University Press, 1990.
[29] Aurélien Bénel. Modéliser ce qui résiste à la modélisation : De la sémantique à la sémiotique. *Revue Ouverte d'Intelligence Artificielle 1* (1), 2020. pp.71-88. ⟨10.5802/roia.4⟩.
[30] Philippe Lacour, Aurélien Bénel, op. cit.